\documentclass[runningheads]{llncs}
\usepackage{graphicx}

\usepackage{tikz}
\usepackage{comment}
\usepackage{amsmath,amssymb} 
\usepackage{color}

\usepackage{amsfonts}
\usepackage{booktabs}
\usepackage{siunitx,booktabs,caption}
\usepackage[accsupp]{axessibility}  

\usepackage[width=122mm,left=12mm,paperwidth=146mm,height=193mm,top=12mm,paperheight=217mm]{geometry}

\usepackage{multirow}
\usepackage{multicol}
\usepackage{makecell}
\usepackage{booktabs} 
\newcommand{\etal}{\textit{et al.}}
\usepackage{pifont}

\usepackage[pagebackref=true,breaklinks=true,letterpaper=true,colorlinks,bookmarks=false]{hyperref}

\begin{document}
\pagestyle{headings}
\mainmatter
\def\ECCVSubNumber{116}  

\title{Towards Grand Unification of Object Tracking} 

\titlerunning{Towards Grand Unification of Object Tracking}
%
\author{Bin Yan$^{1}$\thanks{This work was performed while Bin Yan worked as an intern at
ByteDance. Email: \href{mailto:yan_bin@mail.dlut.edu.cn}{\color{black}{yan\_bin@mail.dlut.edu.cn}.}} \and
Yi Jiang$^{2, \dagger}$ \and
Peize Sun$^{3}$ \and 
Dong Wang$^{1, \dagger}$\and \\
Zehuan Yuan$^{2}$\and
Ping Luo$^{3}$\and
Huchuan Lu$^{1,4}$}
\authorrunning{B. Yan et al.}
%
\institute{$^{1}$ School of Information and Communication Engineering, Dalian University of
Technology, China 
$^{2}$ ByteDance $^{3}$ The University of Hong Kong \\$^{4}$ Peng Cheng Laboratory}
\maketitle

\begin{abstract}

We present a unified method, termed Unicorn, that can simultaneously solve four tracking problems (SOT, MOT, VOS, MOTS) with a single network using the same model parameters. Due to the fragmented definitions of the object tracking problem itself, most existing trackers are developed to address a single or part of tasks and over-specialize on the characteristics of specific tasks. By contrast, Unicorn provides a unified solution, adopting the same input, backbone, embedding, and head across all tracking tasks. For the first time, we accomplish the great unification of the tracking network architecture and learning paradigm. Unicorn performs on-par or better than its task-specific counterparts in 8 tracking datasets, including LaSOT, TrackingNet, MOT17, BDD100K, DAVIS16-17, MOTS20, and BDD100K MOTS. We believe that Unicorn will serve as a solid step towards the general vision model. Code is available at \href{https://github.com/MasterBin-IIAU/Unicorn}{https://github.com/MasterBin-IIAU/Unicorn}.

\keywords{object tracking}
\end{abstract}

\newcommand\blfootnote[1]{%
\begingroup 
\renewcommand\thefootnote{}\footnote{#1}%
\addtocounter{footnote}{-1}%
\endgroup 
}
{
	\blfootnote{
	 ~$^\dagger$ Corresponding authors: \href{mailto:jiangyi.enjoy@bytedance.com}{\color{black}{jiangyi.enjoy@bytedance.com}}, \href{mailto:wdice@dlut.edu.cn}{\color{black}{wdice@dlut.edu.cn}}.
	}
}

\vspace{-8mm}
\section{Introduction}

Compared with weak AI designed for solving one specific task, artificial general intelligence (AGI) is expected to understand or learn any intellectual task that a human being can. Although there is still a large gap between this ambitious goal and the intellectual algorithms of today, some recent works~\cite{MuST,INTERN,florence,Omnivore} have begun to explore the possibility of building general vision models to address several vision tasks simultaneously.

Object tracking is one of the fundamental tasks in computer vision, which aims to build pixel-level or instance-level correspondence between frames and to output trajectories typically in the forms of boxes or masks. Over the years, according to different application scenarios, the object tracking problem has been mainly divided into four separate sub-tasks: Single Object Tracking (SOT)~\cite{LaSOT,trackingnet}, Multiple Object Tracking (MOT)~\cite{MOT17,BDD100K}, Video Object Segmentation (VOS)~\cite{DAVIS17}, and Multi-Object Tracking and Segmentation (MOTS)~\cite{MOTS,BDD100K}. As a result, most tracking approaches are developed for only one of or part of the sub-tasks. Despite convenience for specific applications, this fragmented situation brings into the following drawbacks: (1) Trackers may over-specialize on the characteristic of specific sub-tasks, lacking in the generalization ability. (2) Independent model designs cause redundant parameters. For example, recent deep-learning-based trackers usually adopt similar backbones architectures, but the separate design philosophy hinders the potential reuse of parameters. 
It is natural to ask a question: Can all main-stream tracking tasks be solved by a unified model?

Although some works~\cite{SiamMask,D3S,SiamRCNN,Trades,Trackformer} attempt to unify SOT\&VOS or MOT\& MOTS by adding a mask branch to the existing box-level tracking system, there is still little progress towards the unification of SOT and MOT. There are mainly three obstacles hindering this process. 
(1) The characteristics of tracked objects vary. MOT usually tracks tens even hundreds of instances of specific categories. In contrast, SOT needs to track one target given in the reference frame no matter what class it belongs to. 
(2) SOT and MOT require different types of correspondence.
SOT requires distinguishing the target from the background. However, MOT needs to match the currently detected objects with previous trajectories.
(3) Most SOT methods~\cite{SiameseFC,SiamRPNplusplus,ATOM,DiMP,TransT,STARK} only take a small search region as the input to save computation and filter potential distractors. However, MOT algorithms~\cite{Tracktor,DeepMOT,mot_solver,JDE,FairMOT,CenterTrack,Trackformer} usually take the high-resolution full image as the input for detecting instances as completely as possible.

To conquer these challenges, we propose two core designs: the target prior and the pixel-wise correspondence. To be specific, (1) the target prior is an additional input for the detection head and serves as the switch among four tasks. For SOT\&VOS, the target prior is the propagated reference target map, enabling the head to focus on the tracked target. For MOT\&MOTS, by setting the target prior as zero, the head degenerates into the usual class-specific detection head smoothly.  (2) The pixel-wise correspondence is the similarity between all pairs of points from the reference frame and the current frame. Both the SOT correspondence ($\mathbf{C}^{\mathrm{SOT}}\in\mathbb{R}^{h'w'\times{hw}}$) and the MOT correspondence ($\mathbf{C}^{\mathrm{MOT}}\in\mathbb{R}^{M\times{N}}$) are subsets of the pixel-wise correspondence ($\mathbf{C}_{\mathrm{pix}}\in\mathbb{R}^{hw\times{hw}}$). (3) With the help of the informative target prior and the accurate pixel-wise correspondence, the design of the search region becomes unnecessary for SOT, leading to unified inputs as the full image for SOT and MOT.

Towards the \textbf{uni}fi\textbf{c}ation of \textbf{o}bject t\textbf{r}acki\textbf{n}g, we propose Unicorn, a single network architecture to solve four tracking tasks.
It takes the reference frame and the current frame as the inputs and produces their visual features by a weight-shared backbone. Then a feature interaction module is exploited to build pixel-wise correspondence between two frames. Based on the correspondence, a target prior is generated by propagating the reference target to the current frame. Finally, the target prior and the visual features are fused and sent to the detection head to get the tracked objects for all tasks.

With the unified network architecture, Unicorn can learn from various sources of tracking data and address four tracking tasks with the same model parameters. Extensive experiments show that Unicorn performs on-par or better than task-specific counterparts on 8 challenging benchmarks from four tracking tasks.

We summarize that our work has the following contributions: 
\begin{itemize}
\vspace{-1mm}
	\item{For the first time, Unicorn accomplishes the great unification of the network architecture and the learning paradigm for four tracking tasks.}
	\item{Unicorn bridges the gap among methods of four tracking tasks by the target prior and the pixel-wise correspondence.}
	\item{Unicorn puts forwards new state-of-the-art performance on 8 challenging tracking benchmarks with the same model parameters. This achievement will serve as a solid step towards the general vision model.}
\end{itemize}

\section{Related Work}
\subsection{Task-specific Trackers}

SOT typically specifies one tracked target with a bounding box on the first frame, then requires trackers to predict boxes for the tracked target in the following frames. Considering the uniqueness and the motion continuity of the tracked target, most of the algorithms in SOT~\cite{SiameseFC,SiamRPNplusplus,SiamFC++,ATOM,DiMP,TransT,STARK} track on a small search region rather than the whole image to reduce  computation and to filter distractors. Although achieving great success in the SOT field, search-region-based trackers suffer from the following drawbacks: (1) Due to the limited visual field, it is difficult for these methods to recover from temporary tracking failure, especially in the long-term tracking scenarios. (2) The speed of these methods drops drastically as the number of tracked instances increases. The inefficiency problem restricts the application of SOT trackers in scenarios such as MOT, where there are tens or hundreds of targets to track. To overcome the first problem, some works~\cite{GlobalTrack,SiamRCNN} propose a global-detection-based tracking paradigm. However, these methods either require large modifications to the original detection architecture to integrate the target information or rely on complicated dynamic programming to pick the best tracklet. Besides, both Global-Track~\cite{GlobalTrack} and Siam R-CNN~\cite{GlobalTrack} are developed on two-stage Faster R-CNN, whose detection pipeline is tedious and relies on hand-crafted anchors and ROI-Align. By contrast, in this work, we build our method based on a one-stage, anchor-free detector~\cite{YOLOX}. Furthermore, we demonstrate that only with minimal change to the original detector architecture, we could transform an object detector into a powerful SOT tracker. 

Different from SOT, MOT does not have any given prior on the first frame. Trackers of MOT are required to find and associate all instances of specific classes by themselves. The mainstream methods~\cite{JDE,FairMOT,CenterTrack,QDTrack,TransTrack} follow the tracking-by-detection paradigm. Specifically, an MOT system typically has two main components, an object detector and a certain association strategy. Commonly used detectors include Faster R-CNN~\cite{FasterRCNN}, the YOLO series~\cite{YOLOv3,YOLOX}, CenterNet~\cite{CenterNet}, Sparse R-CNN\cite{sparsercnn}, and Deformable DETR~\cite{DeformableDETR}, etc. Popular association methods include IoU matching~\cite{SORT,TransTrack}, Kalman Filter~\cite{SORT,JDE,FairMOT}, ReID embedding~\cite{DeepSORT,QDTrack,JDE,FairMOT}, Transformer~\cite{TransTrack,Trackformer,MOTR}, or the combination of them\cite{bytetrack}. Although there are some works~\cite{DMAN,FAMNet} introducing SOT trackers for the association, these SOT trackers~\cite{ECO,SiameseFC} are completely independent with the MOT networks, without any weight sharing. There is still a large gap between methods of SOT and MOT.

The goal of VOS is to predict masks for the tracked instances based on the high-quality mask annotations of the first frame. This field is now dominated by memory-network-based methods~\cite{STM,CFBI,STCN}. Although achieving great performance, these methods suffer from the following disadvantages: (1) The memory network brings huge time and space complexity, especially when dealing with high spatial resolution and the long sequence. While these scenarios are quite common in sequences of SOT and MOT. Specifically, the long-term tracking benchmarks~\cite{LaSOT,OxUvA} in SOT usually have thousands of frames per sequence, being more than 20x longer than DAVIS~\cite{DAVIS17}. Meanwhile, the image size in MOT~\cite{BDD100K} can reach 720x1280, while the image size of DAVIS is usually only 480x854. (2) SOTA methods assume that there are always high-quality mask annotations on the first frame. However, high-quality masks demand expensive labor costs and are usually unavailable in real-world applications. To overcome this problem, some works~\cite{SiamMask,D3S,SiamRCNN} attempt to develop weakly-annotated VOS algorithms, which only require box annotation on the first frame. 

MOTS is highly related to MOT by changing the form of boxes to fine-grained representation of masks. MOTS benchmarks~\cite{MOTS,BDD100K} are typically from the same scenarios as those of MOT~\cite{MOT17,BDD100K}. Besides, many MOTS methods are developed upon MOT trackers. Representative approaches include 3D-convolution-based Track R-CNN~\cite{MOTS} and Stem-Seg~\cite{stemseg}, Transformer-based TrackFormer~\cite{Trackformer}, tracking-assisting-detection Trades~\cite{Trades} and Prototype-based PCAN~\cite{PCAN}. 

\subsection{General Vision Models}
Despite the great success of specialized models for diverse tasks, there is still a large gap between the current AI with human-like, omnipotent Artificial General Intelligence (AGI). An important step towards this grand goal is to build a generalist model supporting a broad range of AI tasks. Recent pioneering works~\cite{MuST,INTERN,florence,Omnivore} attempt to approach this goal from different perspectives. Specifically, MuST~\cite{MuST} introduces a multi-task self-training pipeline, which harnesses the knowledge in independent specialized teacher models to train a single general student model. INTERN~\cite{INTERN} proposes a new learning paradigm, which learns with supervisory signals from multiple sources in multiple stages. The developed general vision model generalizes well to different tasks but also has lower requirements on downstream data. Florence~\cite{florence} is a new computer vision foundation model, which expands the representations to different tasks along space, time, and modality. Florence has great transferability and achieves new SOTA results on a wide range of vision benchmarks. OMNIVORE~\cite{Omnivore} proposes a modality-agnostic model which can classify images, videos, and single-view 3D data using the same model parameters.

\begin{figure}[!t]
  \begin{center}
\includegraphics[width=1.0\linewidth]{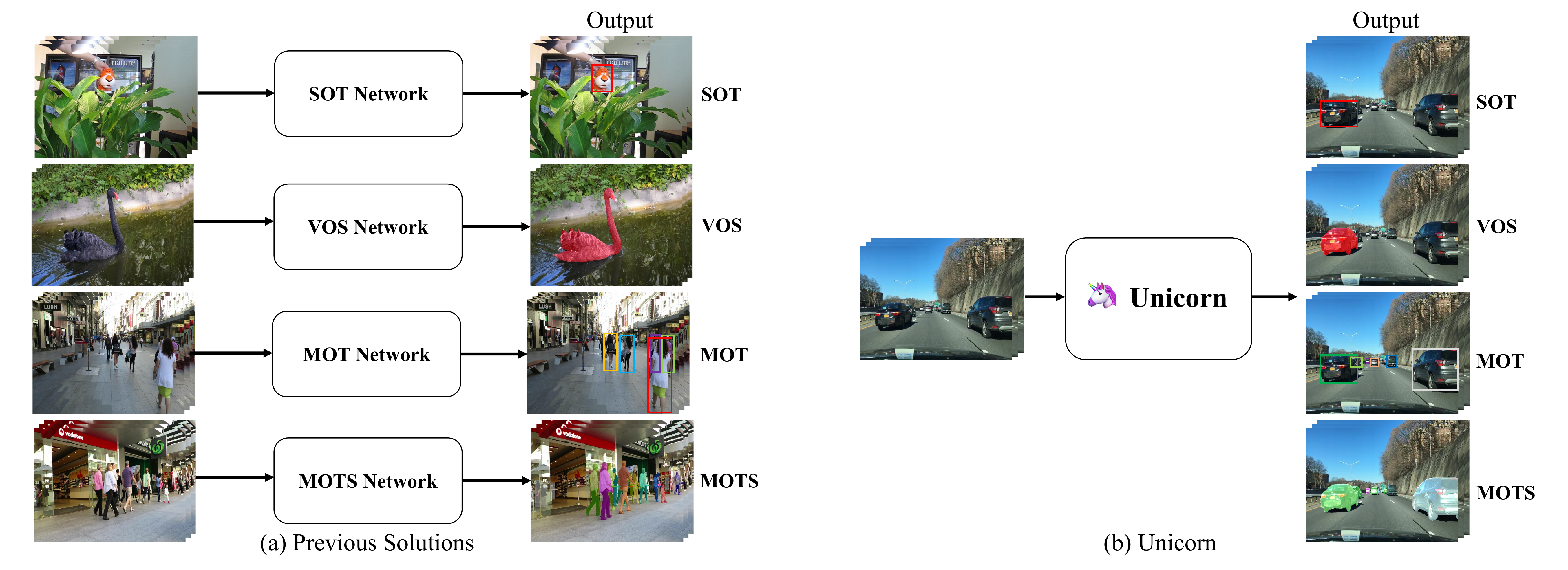}
  \end{center}
  \vspace{-5mm}
  \caption{Comparison between previous solutions and Unicorn.} 
  \label{fig-page1}
\vspace{-3mm}
\end{figure}

\subsection{Unification in Object Tracking}
In the literature, some works~\cite{SiamMask,Trades,uniTrack} attempted to design a unified framework for supporting multiple tracking tasks. Specifically, SiamMask~\cite{SiamMask} is the first work to address SOT and VOS simultaneously. Similarly, TraDes~\cite{Trades} can solve both MOT and MOTS by introducing an extra mask head. Besides, UniTrack~\cite{uniTrack} proposes a high-level tracking framework, which consists of a shared appearance model and a series of unshared tracking heads. It demonstrates that different tracking tasks can share one appearance model for either propagation or association. However, the large discrepancy in tracking heads hinders it from exploiting a large amount of tracking data. Consequently, its performance lags far behind that of SOTA task-specific methods. Moreover, when used for MOT or MOTS, UniTrack requires extra, independent object detectors to provide observation. The extra object detector and the appearance model do not share the same backbone, bringing heavy burdens in parameters. By contrast, Unicorn solves four tracking tasks with one unified network with the same parameters. 
Besides, Unicorn can learn powerful representation from a large amount of labeled tracking data, achieving superior performance on 8 challenging benchmarks. Figure~\ref{fig-page1} shows the comparison between task-specific methods and Unicorn.

\subsection{Correspondence Learning}
Learning accurate correspondence is the key to many vision tasks, such as optical flow~\cite{RAFT}, video object segmentation~\cite{TVOS,CRW}, geometric matching~\cite{GOCor,PDCNet}, etc. The dense correspondence is usually obtained by computing correlation between the embedding maps of two frames. Most existing methods~\cite{RAFT,TVOS,CRW} obtain the embedding maps without considering the information exchange between two images. This could lead to ambiguous or wrong matching when there are many similar patterns or instances on the input images. Although some works~\cite{GOCor,PDCNet} attempt to relieve this problem, they usually require complex optimization or uncertainty modeling. Different from the local comparison, Transformer~\cite{transformer} and its variants~\cite{DeformableDETR} exploit the attention mechanism to capture the long-range dependency within the input sequence. In this work, we demonstrate that these operations can help to learn precise correspondence in object tracking.

\section{Approach}

\begin{figure}[!t]
  \begin{center}
\includegraphics[width=1.0\linewidth]{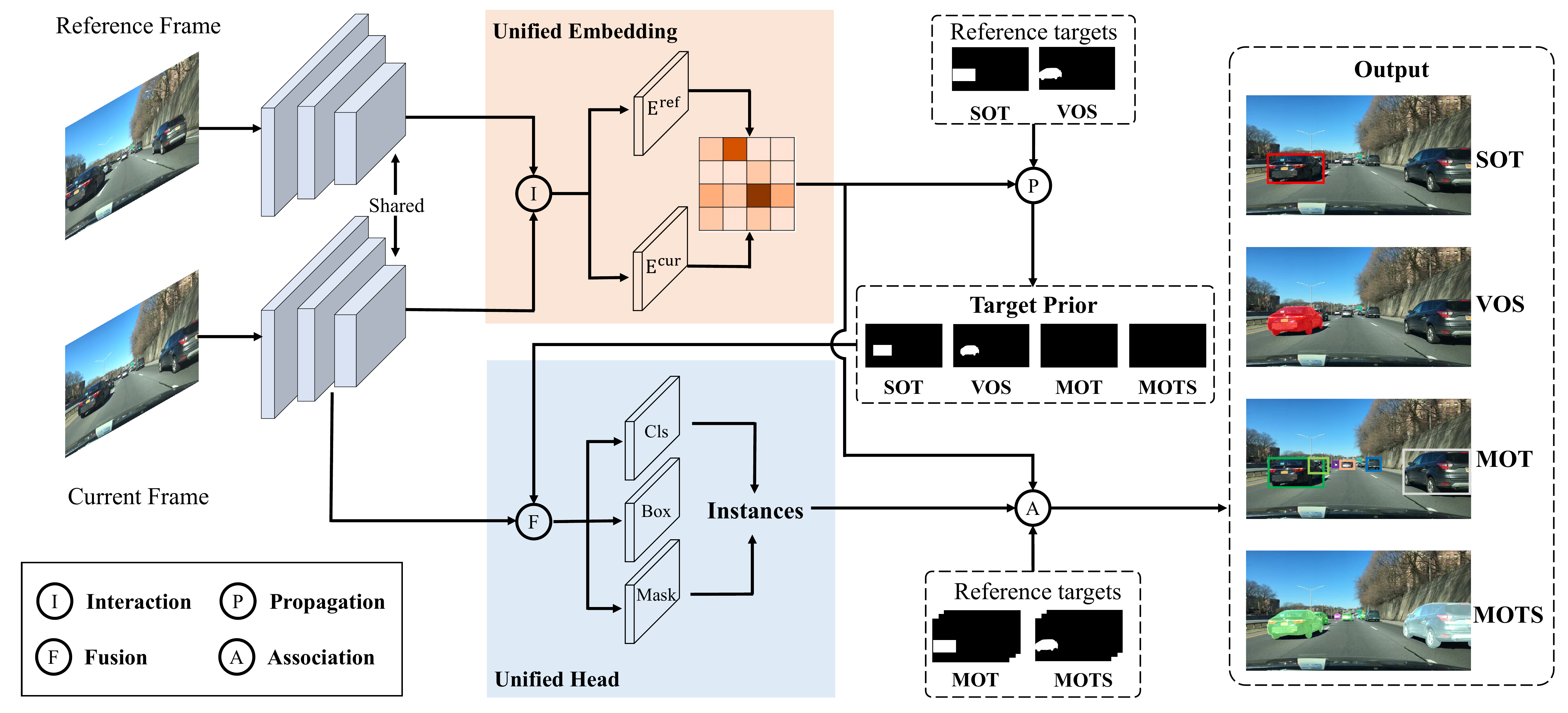}
  \end{center}
  \vspace{-5mm}
  \caption{Unicorn consists of three main components: (1) Unified inputs and backbone (2) Unified embedding (3) Unified head.} 
  \label{fig-framework}
\vspace{-3mm}
\end{figure}

We propose a unified solution for object tracking, called Unicorn, which consists of three main components: unified inputs and backbone; unified embedding and unified head. Three components are responsible for obtaining powerful visual representation, building precise correspondence and detecting diverse tracked targets respectively. The framework of Unicorn is demonstrated in Figure~\ref{fig-framework}. Given the reference frame $\mathbf{I}_{\mathrm{ref}}$, the current frame $\mathbf{I}_{\mathrm{cur}}$, and the reference targets, Unicorn aims at predicting the states of the tracked targets on the current frame for four tasks with a unified network.

\vspace{-5mm}
\subsection{Unified Inputs and Backbone}
For efficiently localizing multiple potential targets, Unicorn takes the whole image (for both the reference frame and the current frame) instead of local search regions as the inputs. This also endows Unicorn high resistance to tracking failure and the ability to re-detect tracked target after disappearance.

During the feature extraction, the reference frame and the current frame are passed through a weight-sharing backbone to get feature pyramid representations(FPN)~\cite{FPN}. To maintain important details and reduce the computational burden during computing correspondence, we choose the feature map with stride 16 as the input of the following embedding module. The corresponding features from the reference and the current frame are termed $\mathbf{F}_{\mathrm{ref}}$ and $\mathbf{F}_{\mathrm{cur}}$ respectively.

\vspace{-5mm}
\subsection{Unified Embedding}
The core task of object tracking is to build accurate correspondence between frames in a video. For SOT and VOS, pixel-wise correspondence propagates the user-provided target from the reference frame (usually the $1^{st}$ frame) to the $t^{th}$ frame, providing strong prior information for the final box or mask prediction. Besides, for MOT and MOTS, instance-level correspondence helps to associate the detected instances on the $t^{th}$ frame to the existing trajectories on the reference frame (usually the ${t-1}^{th}$ frame). 

In Unicorn, given the spatially flattened reference frame embedding $\mathbf{E}_{\mathrm{ref}}\in\mathbb{R}^{hw\times{c}}$ and the current frame embedding $\mathbf{E}_{\mathrm{cur}}\in\mathbb{R}^{hw\times{c}}$, pixel-wise correspondence $\mathbf{C}_{\mathrm{pix}}\in\mathbb{R}^{hw\times{hw}}$ is computed by the matrix multiplication between them. 
For SOT\&VOS taking the full image as the inputs, the correspondence is the the pixel-wise correspondence itself.
For MOT\&MOTS, assume that there are $M$ trajectories on the reference frame and $N$ detected instances on the current frame respectively, the instance-level correspondence  $\mathbf{C}_{\mathrm{inst}}\in\mathbb{R}^{N\times{M}}$ is the matrix multiplication of the reference instance embedding $\mathbf{e}_{\mathrm{ref}}\in\mathbb{R}^{M\times{c}}$ and the current instance embedding $\mathbf{e}_{\mathrm{cur}}\in\mathbb{R}^{N\times{c}}$. The instance embedding $\mathbf{e}$ is extracted from the frame embedding $\mathbf{E}$, where the center of the instance is located.

\begin{equation}
\begin{matrix}\mathbf{C}_{\mathrm{pix}}=\mathrm{softmax}(\mathbf{E}_\mathrm{{cur}}{\mathbf{E}_{\mathrm{ref}}}^T)\\
\mathbf{C}_{\mathrm{inst}}=\mathrm{softmax}(\mathbf{e}_{\mathrm{cur}}{\mathbf{e}_{\mathrm{ref}}}^T)\\\end{matrix}
\end{equation}



\noindent It can be seen that the instance-level correspondence $\mathbf{C}_{\mathrm{inst}}$ required by MOT and MOTS is the sub-matrix of the pixel-wise correspondence $\mathbf{C}_{\mathrm{pix}}$. 
Besides, learning highly discriminative embedding $\{\mathbf{E}_{\mathrm{ref}}, \mathbf{E}_{\mathrm{cur}}\}$ is the key to building precise correspondence for all tracking tasks.

\subsubsection{Feature Interaction.} Due to its advantages of capturing long-range dependency, Transformer~\cite{transformer} is an intuitive choice to enhance the original feature representation $\{\mathbf{F}_{\mathrm{ref}}, \mathbf{F}_{\mathrm{cur}}\}$. However, this could lead to huge memory cost when dealing with high-resolution feature maps, because the memory consumption increases with the length of the input sequence quadratically. To alleviate this problem, we replace the full attention with more memory-efficient deformable attention~\cite{DeformableDETR}. For more accurate correspondence, the enhanced feature maps are upsampled by $2\times$ to obtain high-resolution embeddings on the stride of 8. 

\begin{equation}
    \{\mathbf{E}_{\mathrm{ref}}, \mathbf{E}_{\mathrm{cur}}\}=\mathrm{Upsample}(\mathrm{Attention}(\mathbf{F}_{\mathrm{ref}},\mathbf{F}_{\mathrm{cur}}))
\end{equation}

\subsubsection{Loss.}\label{sec:embed_train}
Ideal embedding should work well on both propagation(SOT, VOS) and association(MOT, MOTS). For SOT\&VOS, although there is no human-annotated label for dense correspondence between frames, the embedding can be supervised by the difference between the propagated result $\mathbf{\widetilde{T}}_{\mathrm{cur}}$ and the ground-truth target map $\mathbf{T}_{\mathrm{cur}}$. Specifically, the shape of target map $\mathbf{T}$ is $hw\times{1}$. The regions where the tracked target exists are equal to one and the other regions are equal to zero. During the propagation, the pixel-wise correspondence $\mathbf{C}_{\mathrm{pix}}$ transforms the reference target map $\mathbf{T}_{\mathrm{ref}}$ to the estimation of the current target map $\mathbf{\widetilde{T}}_{\mathrm{cur}}$.

\begin{equation}
\mathbf{\widetilde{T}}_{\mathrm{cur}}(i,j) = \sum_k \mathbf{C}_{\mathrm{pix}}(i,k) \cdot \mathbf{T}_{\mathrm{ref}}(k,j)
\end{equation}
Besides, for MOT and MOTS, the instance-level correspondence can be learned with standard contrastive learning paradigm. Specifically, assume that the instance $i$ from the current frame is matched with the instance $j$ from the reference frame, then the corresponding ground-truth matrix $\mathbf{G}$ should satisfies that  

\begin{equation}
    \mathbf{G}_{i,k}=\left\{\begin{matrix}0&\ \ k\neq j\\1&\ \ k=j\\\end{matrix}\right.
\end{equation}
Finally, the unified embedding can be optimized end-to-end by Dice Loss~\cite{DiceLoss} for SOT\&VOS or Cross-Entropy Loss for MOT\&MOTS.

\begin{equation}
\mathbf{L}_{\mathrm{corr}}=\left\{\begin{matrix}\mathrm{Dice}(\widetilde{\mathbf{T}}_{\mathrm{cur}},\mathbf{T}_{\mathrm{cur}})&\mathrm{task\ in\ \{SOT,VOS\}}\\\mathrm{CrossEntropy}(\mathbf{C}_{\mathrm{inst}},\mathbf{G})&\ \ \ \mathrm{task\ in\ \{MOT,MOTS\}}\\\end{matrix}\right.
\end{equation}

\subsection{Unified Head}
To achieve the grand unification of object tracking, another important and challenging problem is designing a unified head for four tracking tasks. Specifically, MOT shall detect objects of specific categories. However, SOT needs to detect any target given in the reference frame. To bridge this gap, Unicorn introduces an extra input (called target prior) to the original detector head~\cite{YOLOX,CondInst}. Without any further modification, Unicorn can easily detect various objects needed for four tasks with this unified head. More details about the head architecture can be found in the supplementary materials.


\subsubsection{Target Prior.}
As mentioned in Sec.~\ref{sec:embed_train}, given the reference target map $\mathbf{T}_{\mathrm{ref}}$, the propagated target map $\mathbf{\widetilde{T}}_{\mathrm{cur}}$ can provide strong prior information about the state of the tracked target. This motivates us to take it as a target prior when detecting targets for SOT\&VOS. To be compatible with the original input of the detection head, we first reshape it to $h\times{w}\times{1}$ (i.e. $\mathbf{\widetilde{T}}^{\mathrm{reshape}}_{\mathrm{cur}}\in\mathbb{R}^{h\times{w}\times{1}})$. Meanwhile, when dealing with MOT\&MOTS, we can simply set this prior to zero. Formally, the target prior $\mathbf{P}$ satisfies that

\begin{equation}
\mathbf{P}=\left\{\begin{matrix}\mathbf{\widetilde{T}}^{\mathrm{reshape}}_{\mathrm{cur}}&\mathrm{task\ in\ \{SOT,VOS\}}\\\mathbf{0}&\ \ \ \mathrm{task\ in\ \{MOT,MOTS\}}\\\end{matrix}\right.
\end{equation}

\vspace{-5mm}
\subsubsection{Feature Fusion.}
The unified head takes the original FPN feature $\mathbf{F}\in\mathbb{R}^{h\times{w}\times{c}}$ and the target prior $\mathbf{P}\in\mathbb{R}^{h\times{w}\times{1}}$ as the inputs. Unicorn fuses these two inputs with broadcast sum and passes the fused feature $\mathbf{F}^{'}\in\mathbb{R}^{h\times{w}\times{c}}$ to the original detection head. This fusion strategy has the following advantages. (1) The fused features are seamlessly compatible with four tasks. Specifically, for MOT\&MOTS, the target prior is equal to zero. Then the fused feature $\mathbf{F}^{'}$ degenerates back to the original FPN feature $\mathbf{F}$ to detect objects of specific classes. For SOT\&VOS, the target prior with strong target information can enhance the original FPN feature and makes the network focus on the tracked target.(2) The architecture is simple, without introducing complex changes to the original detection head. Furthermore, the consistent architecture also enables Unicorn to fully exploit the pretrained weights of the original object detector.

\subsection{Training and Inference}
\subsubsection{Training.}
The whole training process divides into two stages: SOT-MOT joint training and VOS-MOTS joint training. In the first stage, the network is end-to-end optimized with the correspondence loss and the detection loss using data from SOT\&MOT. In the second stage, a mask branch is added and optimized with the mask loss using data from VOS\&MOTS with other parameters fixed. 
\vspace{-2mm}
\subsubsection{Inference.}
During the test phase, for SOT\&VOS, the reference target map is generated once on the first frame and kept fixed in the following frames. Unicorn directly picks the box or mask with the highest confidence score as the final tracking result, without any hyperparameter-sensitive post-processing like cosine window. Besides, Unicorn only needs to run the heavy backbone and the correspondence once, while running the lightweight head rather than the whole network $N$ times, leading to higher efficiency. For MOT\&MOTS, Unicorn detects all objects of the given categories and simultaneously outputs corresponding instance embeddings. 
The later association is performed based on the embeddings and the motion model for BDD100K and MOT17 respectively.

\section{Experiments}
\subsection{Implementation Details}
When comparing with state-of-the-art methods, we choose ConvNeXt-Large~\cite{ConvNeXt} as the backbone. In ablations, we report the results of our method with ConvNeXt-Tiny~\cite{ConvNeXt} and ResNet-50~\cite{ResNet} as the backbone. The input image size is $800\times1280$ and the shortest side ranges from 736 to 864 during multi-scale training. The model is trained on 16 NVIDIA Tesla A100 GPU with a global batch size of 32. To avoid inaccurate statistics estimation, we replace all Batch Normalization~\cite{BatchNormalization} with Group Normalization~\cite{GroupNormalization}. Two training stages randomly sample data from SOT\&MOT datasets and VOS\&MOTS datasets, respectively. Each training stage consists of 15 epochs with 200,000 pairs of frames in every epoch. The optimizer is Adam-W~\cite{AdamW} with weight decay of $5e^{-4}$ and momentum of 0.9. The initial learning rate is $2.5e^{-4}$ with 1 epoch warm-up and the cosine annealing schedule. More details can be found in the supplementary materials.

In Section \ref{exp_sot}-\ref{exp_mots}, we compare Unicorn with task-specific counterparts in 8 tracking datasets. In each benchmark, the \textcolor[rgb]{1,0,0}{\textbf{red bold}} font and the \textcolor[rgb]{0,0,1}{blue} font indicate the best two results. 
Unicorn in four tasks uses the same model parameters.

\subsection{Evaluations on Single Object Tracking}
\label{exp_sot}

We compare Unicorn with state-of-the-art SOT trackers on two popular and challenging benchmarks, LaSOT~\cite{LaSOT} and TrackingNet~\cite{trackingnet}. Both datasets evaluate the tracking performance with the following measures: Success, precision($P$) and normalized precision($P_{norm}$). All these measures are the higher the better. 

\textbf{LaSOT}. LaSOT~\cite{LaSOT} is a large-scale long-term tracking
benchmark, which contains 280 videos in the test set with an average length of 2448 frames. Tab~\ref{tab-sot} shows that Unicorn achieves new state-of-the-art Success and Precision of 68.5\% and 74.1\% respectively. It is also worth noting that Unicorn surpasses the previous best global-detection-based tracker Siam R-CNN~\cite{SiamRCNN} by a large margin (68.5\% vs 64.8\%) with a much simpler network architecture and tracking strategy (directly picking the top-1 vs tracklet dynamic programming). 

\textbf{TrackingNet}. TrackingNet~\cite{trackingnet} is a large-scale short-term
tracking benchmark containing 511 videos in the test set. As reported in Tab~\ref{tab-sot}, Unicorn surpasses all previous methods with a Success of 83.0\% and a Precision of 82.2\%. 

\begin{table*}[!t]
\caption{State-of-the-art comparison on LaSOT~\cite{LaSOT} and TrackingNet~\cite{trackingnet}.}
\label{tab-sot}
\vspace{-3mm}
\begin{center}
{\resizebox{0.85\linewidth}{!}{
\begin{tabular}{c|c|ccc|ccc}
\hline
\multirow{2}{*}{Method} & \multirow{2}{*}{Source} &\multicolumn{3}{c|}{LaSOT~\cite{LaSOT}}	
&\multicolumn{3}{c}{TrackingNet~\cite{trackingnet}}\\
\cline{3-8}
& &Success	&P$_{norm}$	&P	&Success	&P$_{norm}$	&P	\\
\hline
SiamFC~\cite{SiameseFC}&ECCVW2016&33.6&42.0&33.9&57.1&66.3&53.3\\
UniTrack~\cite{uniTrack}&NeurIPS2021&35.1&-&32.6&-&-&-\\
ATOM~\cite{ATOM}	    &CVPR2019	&51.5	&57.6	&50.5	&70.3	&77.1	&64.8	\\
SiamPRN++~\cite{SiamRPNplusplus}&CVPR2019	&49.6	&56.9	&49.1	&73.3	&80.0	&69.4	\\
DiMP~\cite{DiMP}	    &ICCV2019	&56.9	&65.0	&56.7	&74.0	&80.1	&68.7	\\
GlobalTrack~\cite{GlobalTrack} & AAAI2020 &52.1 &- &52.7 &70.4 &75.4 &65.6 \\
SiamFC++~\cite{SiamFC++}	&AAAI2020	&54.4	&62.3	&54.7	&75.4	&80.0	&70.5	\\
D3S~\cite{D3S}	    &CVPR2020	  &-	&-	&-	&72.8	&76.8	&66.4\\
PrDiMP~\cite{PrDiMP}&CVPR2020	&59.8	&68.8	&60.8	&75.8	&81.6	&70.4\\
Siam R-CNN~\cite{SiamRCNN}   &CVPR2020	&64.8   &72.2 &- &81.2 &85.4 &80.0\\
KYS~\cite{KYS}	    &ECCV2020	&55.4	&63.3	&-	&74.0	&80.0	&68.8	\\
Ocean~\cite{Ocean}	&ECCV2020	&56.0	&65.1	&56.6	&-	&-	&-\\
TrDiMP~\cite{TMT} &CVPR2021 &63.9 &- &61.4 &78.4 &83.3 &73.1\\
TransT~\cite{TransT}	&CVPR2021	&64.9	&73.8	&69.0	&81.4	&\textcolor[rgb]{0,0,1}{86.7}	&\textcolor[rgb]{0,0,1}{80.3}	\\
AutoMatch~\cite{Automatch} &ICCV2021 &58.2 &- &59.9 &76.0 &- &72.6\\
SAOT~\cite{SAOT} &ICCV2021 &61.6 &70.8 &- &- &- &-\\
KeepTrack~\cite{KeepTrack} &ICCV2021 &\textcolor[rgb]{0,0,1}{67.1} &\textcolor[rgb]{1,0,0}{\textbf{77.2}} &\textcolor[rgb]{0,0,1}{70.2} &- &- &-\\
STARK~\cite{STARK} &ICCV2021 &\textcolor[rgb]{0,0,1}{67.1} &\textcolor[rgb]{0,0,1}{77.0} &- &\textcolor[rgb]{0,0,1}{82.0} &\textcolor[rgb]{1,0,0}{\textbf{86.9}} &-\\
\textbf{Unicorn} &Ours &\textcolor[rgb]{1,0,0}{\textbf{68.5}} &76.6 &\textcolor[rgb]{1,0,0}{\textbf{74.1}} &\textcolor[rgb]{1,0,0}{\textbf{83.0}} &86.4 &\textcolor[rgb]{1,0,0}{\textbf{82.2}}\\
\hline
\end{tabular}}}
\vspace{-10mm}
\end{center}
\end{table*}

\subsection{Evaluations on Multiple Object Tracking}
\label{exp_mot}
We compare Unicorn with state-of-the-art MOT trackers on two challenging benchmarks: MOT17~\cite{MOT17} and BDD100K~\cite{BDD100K}. The common metrics include Multiple-Object Tracking Accuracy (MOTA), Identity F1 Score (IDF1), False Positives (FP), False Negatives (FN), the percentage of Mostly Tracked Trajectories (MT) and Mostly Lost Trajectories (ML), Identity Switches (IDS). Among them, MOTA is the primary metric to measure the overall detection and tracking performance, IDF1 is used to measure the trajectory identity accuracy.

\begin{table*}[!t]
\caption{State-of-the-art comparison on MOT17~\cite{MOT17} test set.}
\label{tab-mot17}
\vspace{-3mm}
\begin{center}
{\resizebox{0.9\linewidth}{!}{
\begin{tabular}{ l | c c c c c c c c c}
\toprule
Tracker & MOTA$\uparrow$ & IDF1$\uparrow$ & HOTA$\uparrow$ & MT$\uparrow$ & ML$\downarrow$ & FP$\downarrow$ & FN$\downarrow$ & IDs$\downarrow$\\
\midrule
Chained-Tracker \cite{ChainedTrack} & 66.6 & 57.4 & 49.0 & 37.8\% & 18.5\% & 22284 & 160491 & 5529\\
CenterTrack \cite{CenterTrack} & 67.8 & 64.7 & 52.2 & 34.6\% & 24.6\% & \textcolor[rgb]{1,0,0}{\textbf{18498}} & 160332 & 3039\\
QuasiDense \cite{QDTrack} & 68.7 & 66.3 & 53.9 & 40.6\% & 21.9\% & 26589 & 146643 & 3378\\
TraDes \cite{Trades} & 69.1 & 63.9 & 52.7 & 36.4\% & 21.5\% & \textcolor[rgb]{0,0,1}{20892} & 150060 & 3555\\
SOTMOT \cite{SOTMOT} & 71.0 & 71.9 & - & 42.7\% & 15.3\% & 39537 & 118983 & 5184\\
TransCenter \cite{Transcenter} & 73.2 & 62.2 & 54.5 & 40.8\% & 18.5\% & 23112 & 123738 & 4614\\
MOTR \cite{MOTR} & 73.4 & 68.6 & 57.8 & 42.9\% & 19.1\% & 27939 & 119589 & \textcolor[rgb]{0,0,1}{2439}\\
FairMOT \cite{FairMOT} & 73.7 & 72.3 & 59.3 & 43.2\% & 17.3\% & 27507 & 117477 & 3303\\
TrackFormer~\cite{Trackformer} & 74.1 & 68.0 & - & - & - & 34602 & 108777 & 2829\\
CSTrack \cite{CSTrack} & 74.9 & 72.6 & 59.3 & 41.5\% & 17.5\% & 23847 & 114303 & 3567\\
TransTrack \cite{TransTrack} & 75.2 & 63.5 & 54.1 & \textcolor[rgb]{0,0,1}{55.3\%} & \textcolor[rgb]{1,0,0}{\textbf{10.2\%}} & 50157 & \textcolor[rgb]{0,0,1}{86442} & 3603\\
OMC \cite{OMC} & 76.3 & 72.3 & - & 44.8\% & 15.5\% & - & - & -\\
CorrTracker \cite{CorrTrack} & 76.5 & 73.6 & \textcolor[rgb]{0,0,1}{60.7} & 47.6\% & 12.7\% & 29808 & 99510 & 3369\\
TransMOT \cite{TransMOT} & \textcolor[rgb]{0,0,1}{76.7} & \textcolor[rgb]{0,0,1}{75.1} & \textcolor[rgb]{1,0,0}{\textbf{61.7}} & 51.0\% & 16.4\% & 36231 & 93150 & \textcolor[rgb]{1,0,0}{\textbf{2346}} \\
\textbf{Unicorn} & \textcolor[rgb]{1,0,0}{\textbf{77.2}} & \textcolor[rgb]{1,0,0}{\textbf{75.5}} & \textcolor[rgb]{1,0,0}{\textbf{61.7}} & \textcolor[rgb]{1,0,0}{\textbf{58.7\%}} & \textcolor[rgb]{0,0,1}{11.2\%} & 50087 & \textcolor[rgb]{1,0,0}{\textbf{73349}} & 5379\\
\bottomrule
\end{tabular}}

}
\end{center}
\end{table*}

\begin{table*}[t]
    \caption{State-of-the-art comparison on BDD100K~\cite{BDD100K} tracking validation set.}
    \centering
    {    \resizebox{\linewidth}{!}{
        \begin{tabular}{lccccccccccc}
            \toprule
            Method                  & Split & mMOTA$\uparrow$ & mIDF1$\uparrow$ & MOTA$\uparrow$ & IDF1$\uparrow$ & FN$\downarrow$ & FP$\downarrow$ & ID Sw.$\downarrow$ & MT$\uparrow$  & ML$\downarrow$ & mAP$\uparrow$ \\
            \midrule
            Yu~\etal~\cite{BDD100K} & val   & 25.9             & 44.5             & 56.9            & 66.8            & 122406          & 52372           & 8315                & 8396           & 3795            & 28.1           \\
            QDTrack~\cite{QDTrack}                    & val   & \textcolor[rgb]{0,0,1}{36.6}    & \textcolor[rgb]{0,0,1}{50.8}    & \textcolor[rgb]{0,0,1}{63.5}   & \textcolor[rgb]{1,0,0}{\textbf{71.5}}   & \textcolor[rgb]{0,0,1}{108614} & \textcolor[rgb]{0,0,1}{46621}  & \textcolor[rgb]{1,0,0}{\textbf{6262}}       & \textcolor[rgb]{0,0,1}{9481}  & \textcolor[rgb]{0,0,1}{3034}   & \textcolor[rgb]{0,0,1}{32.6}  \\
            \textbf{Unicorn} & val &\textcolor[rgb]{1,0,0}{\textbf{41.2}}&\textcolor[rgb]{1,0,0}{\textbf{54.0}}&\textcolor[rgb]{1,0,0}{\textbf{66.6}}&\textcolor[rgb]{0,0,1}{71.3}&\textcolor[rgb]{1,0,0}{\textbf{95454}}&\textcolor[rgb]{1,0,0}{\textbf{41648}}&\textcolor[rgb]{0,0,1}{10876}&\textcolor[rgb]{1,0,0}{\textbf{10296}}&\textcolor[rgb]{1,0,0}{\textbf{2505}}&\textcolor[rgb]{1,0,0}{\textbf{41.4}} \\
            \bottomrule
        \end{tabular}
    }

}
    \label{tab:bdd}
\end{table*}

\textbf{MOT17}. The MOT17 focuses on pedestrian tracking and includes 7 sequences in the training set and 7 sequences in the test set. We compare Unicorn with previous methods under the private detection protocol on the test set of MOT17. Tab~\ref{tab-mot17} demonstrates that Unicorn achieves the best MOTA and IDF1, surpassing the previous SOTA method by 0.5\% and 0.4\% respectively. 

\textbf{BDD100K MOT}. BDD100K is a large-scale dataset of visual driving scenes and requires tracking 8 categories of instances. To evaluate the average performance across 8 classes, BDD100K additionally introduces two measures: mMOTA and mIDF1. Different from MOT17, BDD100K is annotated at only 5 FPS. The low frame-rate brings difficulty to motion models commonly used for MOT17. As shown in Tab~\ref{tab:bdd}, Unicorn achieves the best performance, largely surpassing the previous SOTA method QDTrack~\cite{QDTrack} on the val set. Specifically, the improvement is up to 4.6\% and 3.2\% in terms of mMOTA and mIDF1 respectively.

\subsection{Evaluations on Video Object Segmentation}
\label{exp_vos}
We further evaluate the ability of Unicorn to perform VOS on DAVIS~\cite{DAVIS17} 2016 and 2017. Both datasets evaluate methods with the region similarity $\mathcal{J}$, the contour accuracy $\mathcal{F}$, and the average of them $\mathcal{J\&F}$.

\begin{table*}[!t]
\caption{State-of-the-art comparison on the validation set of the DAVIS-2016 and the DAVIS-2017. OL: online learning, Memory: using an external memory bank.}
\centering{}{\footnotesize{}}%
{\resizebox{0.9\linewidth}{!}{
\setlength{\tabcolsep}{2pt}
\begin{tabular}{lccccccccc}
\toprule 
{\footnotesize{}Init}\hspace{-4mm} & {\footnotesize{}Method} & {\footnotesize{}OL} & {\footnotesize{}Memory} & {\footnotesize{}$(\mathcal{J}$\&$\mathcal{F})^{16}$} & {\footnotesize{}$\mathcal{J}^{16}$} & {\footnotesize{}$\mathcal{F}^{16}$} & {\footnotesize{}$(\mathcal{J}$\&$\mathcal{F})^{17}$} & {\footnotesize{}$\mathcal{J}^{17}$} & {\footnotesize{}$\mathcal{F}^{17}$}\tabularnewline

\midrule

\parbox[t]{2mm}{\multirow{11}{*}{\rotatebox[origin=c]{90}{mask}}}
&FAVOS~\cite{FAVOS}&\ding{55}&\ding{55}&81.0&82.4&79.5&58.2&54.6&61.8\tabularnewline
&OSMN~\cite{OSMN}&\ding{55}&\ding{55}&73.5&74.0&72.9&54.8&52.5&57.1\tabularnewline
&VideoMatch~\cite{videomatch}&\ding{55}&\ding{55}&-&81.0&-&56.5&-&-\tabularnewline
&UniTrack~\cite{uniTrack}&\ding{55}&\ding{51}&-&-&-&-&58.4&-\tabularnewline
&RANet~\cite{RANet}&\ding{55}&\ding{55}&85.5&85.5&85.4&65.7&63.2&68.2\tabularnewline
&FRTM~\cite{FRTM}&\ding{51}&\ding{51}&83.5&83.6&83.4&76.7&73.9&79.6\tabularnewline

&TVOS~\cite{TVOS}&\ding{55}&\ding{51}&-&-&-&72.3&69.9&74.7\tabularnewline

&LWL~\cite{LWL}&\ding{51}&\ding{51}&-&-&-&81.6&79.1&84.1\tabularnewline

&STM~\cite{STM}&\ding{55}&\ding{51}&89.3&88.7&89.9&81.8&79.2&84.3\tabularnewline

&CFBI~\cite{CFBI}&\ding{55}&\ding{51}&89.4&88.3&90.5&81.9&79.1&84.6\tabularnewline

&HMMN~\cite{HMMN}&\ding{55}&\ding{51}&90.8&89.6&92.0&84.7&81.9&87.5\tabularnewline

&STCN~\cite{STCN}&\ding{55}&\ding{51}&91.6&90.8&92.5&85.4&82.2&88.6\tabularnewline

\midrule

\parbox{2mm}{\multirow{5}{*}{\rotatebox[origin=c]{90}{bbox}}}
&SiamMask~\cite{SiamMask}&\ding{55}&\ding{55}&69.8&71.7&67.8&56.4&54.3&58.5\tabularnewline

&D3S~\cite{D3S}&\ding{55}&\ding{55}&\textcolor[rgb]{0,0,1}{74.0}&\textcolor[rgb]{0,0,1}{75.4}&\textcolor[rgb]{0,0,1}{72.6}&60.8&57.8&63.8\tabularnewline

&Siam R-CNN~\cite{SiamRCNN}&\ding{55}&\ding{55}&-&-&-&\textcolor[rgb]{1,0,0}{\textbf{70.6}}&\textcolor[rgb]{1,0,0}{\textbf{66.1}}&\textcolor[rgb]{1,0,0}{\textbf{75.0}}\tabularnewline

&\textbf{Unicorn}&\ding{55}&\ding{55}&\textcolor[rgb]{1,0,0}{\textbf{87.4}}&\textcolor[rgb]{1,0,0}{\textbf{86.5}}&\textcolor[rgb]{1,0,0}{\textbf{88.2}}&\textcolor[rgb]{0,0,1}{69.2}&\textcolor[rgb]{0,0,1}{65.2}&\textcolor[rgb]{0,0,1}{73.2}\tabularnewline

\bottomrule
\end{tabular}}}
\label{tab:vos}
\end{table*}

\textbf{DAVIS-16}. DAVIS-16 includes 20 videos in the validation set and there is only one tracked target in each sequence. Tab.~\ref{tab:vos} demonstrates that Unicorn achieves the best results among methods with bounding-box initialization, even surpassing RANet~\cite{RANet} and FRTM~\cite{FRTM} with mask initialization. Meanwhile, Unicorn outperforms its multi-task counterparts SiamMask~\cite{SiamMask} by a large margin of 17.6\% in terms of $\mathcal{J\&F}$.

\textbf{DAVIS-17}. DAVIS-17 contains 30 videos in the validation set and there could be multiple tracked targets in each sequence. As shown in Tab.~\ref{tab:vos}, compared with the previous best box-initialized method Siam R-CNN~\cite{SiamRCNN}, Unicorn achieves competitive results with a much simpler architecture. Specifically, Siam R-CNN~\cite{SiamRCNN} uses an extra Box2Seg network, which is completely independent from the box-based tracker without any weight sharing. However, Unicorn can predict both boxes and masks with a unified head. Although there is still gap between the performance of Unicorn with that of SOTA VOS methods with mask initialization, Unicorn can address four tracking tasks with the same model parameters, while HMMN~\cite{HMMN} and STCN~\cite{STCN} can only be used in the VOS task.

\subsection{Evaluations on Multi-Object Tracking and Segmentation}
\label{exp_mots}

Finally, we evaluate the ability of Unicorn to perform MOTS on MOTS20~\cite{MOTS} and the BDD100K MOTS~\cite{BDD100K}. The main evaluation metrics are sMOTSA and mMOTSA, which are the variants of MOTA and are calculated based on the mask overlap. The other metrics are the same as those in MOT.

\textbf{MOTS20 Challenge}. MOTS20 Challenge has 4 sequences in the train set and 4 sequences in the test set. As shown in Tab.~\ref{tab:mots}, Unicorn achieves state-of-the-art performance, surpassing the second-best method PoinTrackV2~\cite{PointTrackV2} by a large margin of 3.3\% on sMOTSA.

\textbf{BDD100K MOTS Challenge}. BDD100K MOTS Challenge includes 37 sequences in the validation set. Tab.~\ref{tab:bdd_mots} demonstrates that Unicorn outperforms the previous best method PCAN~\cite{PCAN} by a large margin (i.e. mMOTSA +2.2\%, mAP +5.5\%). Meanwhile, Unicorn does not use any complex design like space-time memory or prototypical network as in PCAN, bringing into a simpler pipeline.

\begin{table*}[t]
    \caption{State-of-the-art comparison on the MOTS~\cite{MOTS} test set.}
    \centering
    {    \resizebox{0.9\linewidth}{!}{
        \begin{tabular}{lccccccc}
            \toprule
            Method                  & sMOTSA$\uparrow$ & IDF1$\uparrow$ & MT$\uparrow$  & ML $\downarrow$ & FP$\downarrow$ & FN$\downarrow$ & ID Sw.$\downarrow$ \\
            \midrule
            Track R-CNN~\cite{MOTS}&40.6&42.4&38.7\%&21.6\%&1261&12641&567\\
            TraDeS~\cite{Trades}&50.8&58.7&49.4\%&18.3\%&1474&9169&492\\
            TrackFormer~\cite{Trackformer}&54.9&\textcolor[rgb]{0,0,1}{63.6}&-&-&2233&7195&\textcolor[rgb]{1,0,0}{\textbf{278}}\\
            PointTrackV2~\cite{PointTrackV2}&\textcolor[rgb]{0,0,1}{62.3}&42.9&\textcolor[rgb]{0,0,1}{56.7\%}&\textcolor[rgb]{0,0,1}{12.5\%}&\textcolor[rgb]{1,0,0}{\textbf{963}}&\textcolor[rgb]{0,0,1}{5993}&541\\
            \textbf{Unicorn}&\textcolor[rgb]{1,0,0}{\textbf{65.3}}&\textcolor[rgb]{1,0,0}{\textbf{65.9}}&\textcolor[rgb]{1,0,0}{\textbf{64.9\%}}&\textcolor[rgb]{1,0,0}{\textbf{10.1\%}}&\textcolor[rgb]{0,0,1}{1364}&\textcolor[rgb]{1,0,0}{\textbf{4452}}&\textcolor[rgb]{0,0,1}{398}\\
            \bottomrule
        \end{tabular}
    }
}
    \label{tab:mots}
\end{table*}

\begin{table*}[t]
    \caption{State-of-the-art comparison on the BDD100K MOTS validation set.}
    \centering
    {    \resizebox{0.9\linewidth}{!}{
        \begin{tabular}{lcccccc}
            \toprule
            Method & Online & mMOTSA$\uparrow$ & mMOTSP$\uparrow$ & mIDF1$\uparrow$ & ID Sw.$\downarrow$&mAP$\uparrow$ \\
            \midrule
            SortIoU&\ding{51}&10.3&59.9&21.8&15951&22.2\\
            MaskTrackRCNN~\cite{VIS}&\ding{51}&12.3&59.9&26.2&9116&22.0\\
            STEm-Seg~\cite{stemseg}&\ding{55}&12.2&58.2&25.4&8732&21.8\\
            QDTrack-mots~\cite{QDTrack}&\ding{51}&22.5&59.6&40.8&1340&22.4\\
            QDTrack-mots-fix~\cite{QDTrack}&\ding{51}&23.5&66.3&44.5&\textcolor[rgb]{0,0,1}{973}&25.5\\
            PCAN~\cite{PCAN}&\ding{51}&\textcolor[rgb]{0,0,1}{27.4}&\textcolor[rgb]{0,0,1}{66.7}&\textcolor[rgb]{1,0,0}{\textbf{45.1}}&\textcolor[rgb]{1,0,0}{\textbf{876}}&\textcolor[rgb]{0,0,1}{26.6}\\
            \textbf{Unicorn}&\ding{51}&\textcolor[rgb]{1,0,0}{\textbf{29.6}}&\textcolor[rgb]{1,0,0}{\textbf{67.7}}&\textcolor[rgb]{0,0,1}{44.2}&1731&\textcolor[rgb]{1,0,0}{\textbf{32.1}}\\
            \bottomrule
        \end{tabular}
    }}
    \label{tab:bdd_mots}
\end{table*}

\vspace{-1mm}
\subsection{Ablations and the Other Analysis}

In this section, we conduct component-wise analysis by a series of variants, and make visualization to better understand our method. For the ablations, we choose Unicorn with ConvNeXt-Tiny~\cite{ConvNeXt} backbone as the baseline. The detailed results are demonstrated in Tab.~\ref{table:Ablations}.

\textbf{Backbone}. We implement a variant of Unicorn with ResNet-50~\cite{ResNet} as the backbone. Although the overall performance of this version is lower than the baseline, this variant still achieves superior performance on four tasks.

\textbf{Interaction}. Besides the memory-efficient deformable attention~\cite{DeformableDETR}, we compare the full attention~\cite{transformer} and the convolution operation, which does not exchange information between frames. Experiments show that deformable attention obtains better performance than the full attention, while consuming much less memory. Moreover, the results of the convolution are lower than the baseline, showing the importance of interaction for accurate correspondence.

\textbf{Fusion}. To fuse the FPN features with the target prior, apart from the broadcast sum fusion, we compare other two methods: concatenation, and without the target prior. Experiments show that the performance of SOT and VOS drops significantly after removing the target prior, demonstrating the importance of this design. 
Besides, broadcast sum performs better than concatenation.

\textbf{Single Task}. We compare with training four independent models for different tasks. Experiments show that our unified model performs on-par with independently trained counterparts, while being much more parameter-efficient.  

\textbf{Speed}. 
We develop a light-weight variant with a lower input resolution of 640x1024. Experiments show that the real-time version does not only achieves competitive performance but also can run in real-time at more than 20 FPS.

\begin{figure}[!t]
  \begin{center}
\includegraphics[width=0.8\linewidth]{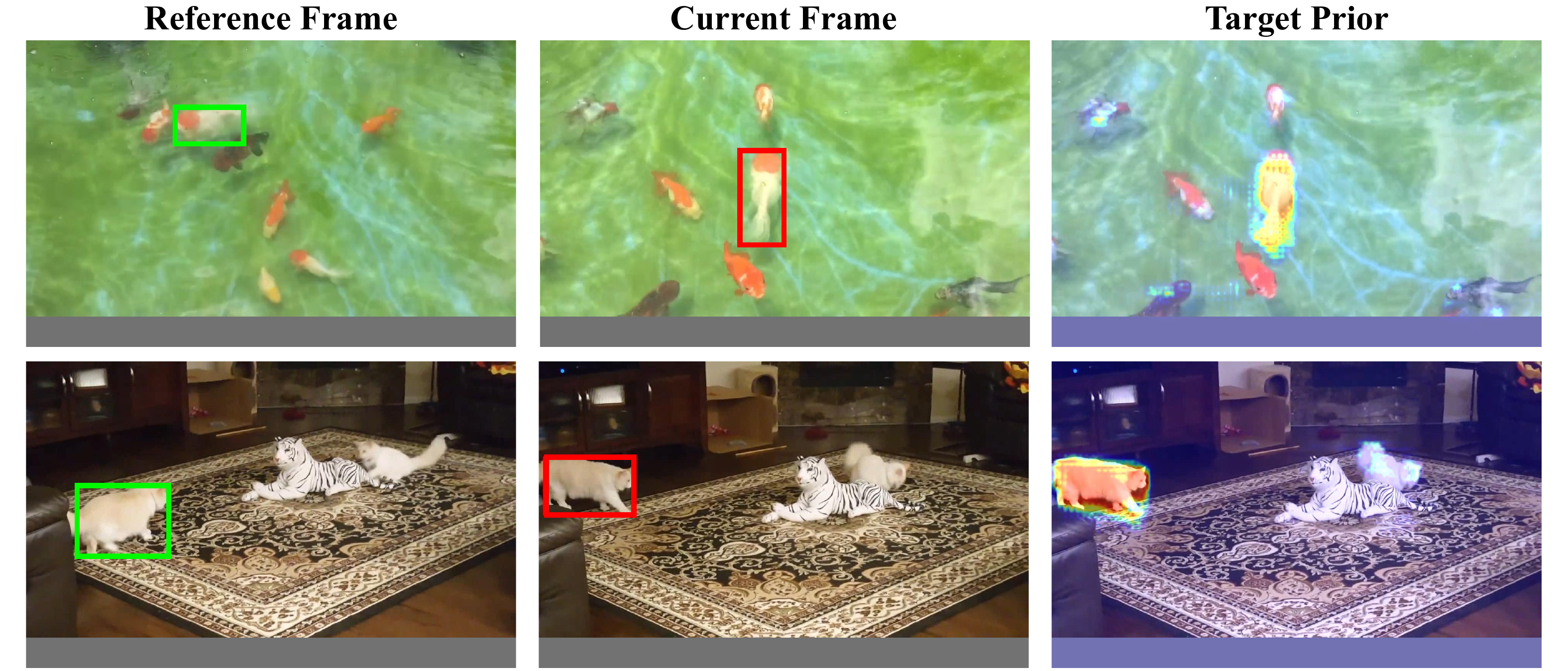}
  \end{center}
  \vspace{-5mm}
  \caption{Visualization of the target prior.} 
  \label{fig-vis}
\vspace{-5mm}
\end{figure}

\textbf{Visualization}. 
In Figure~\ref{fig-vis}, given the tracked target (highlighted with a green box) on the reference frame, we visualize the predicted target prior on the current frame. It can be seen that Unicorn can predict accurate correspondence in challenging scenarios even though there are many similar distractors.

\setlength\tabcolsep{.7em}
\begin{table}[t]
\caption{Ablations and comparisons. Our baseline model are underlined.}
\centering
{\resizebox{1.0\textwidth}{!}{
\begin{tabular}{clccccc}
\toprule
\multirow{2}{*}{Experiment} & \multirow{2}{*}{Method} & \multicolumn{1}{c}{\underline{SOT}} & 
\multicolumn{1}{c}{\underline{MOT}} & 
\multicolumn{1}{c}{\underline{VOS}} & 
\multicolumn{1}{c}{\underline{MOTS}} & 
\multirow{2}{*}{FPS} \\
& &LaSOT(AUC)&BDD(mMOTA)&DAVIS17($\mathcal{J\&F}$)&BDD(mMOTSA)&  \\
\midrule 
\midrule
\multirow{2}{*}{Backbone}  & \underline{ConvNeXt-Tiny} &67.7&39.9&68.0&29.7&14  \\
                            & ResNet-50 &65.3&35.1&66.2&30.8&13 \\ \midrule
\multirow{3}{*}{Interaction}      
    &\underline{Deformable Att}&67.7&39.9&68.0&29.7&14 \\
    & Full Att&67.1&38.5&66.9&26.7&13 \\ 
    & Conv &66.8 &37.6 &66.6&27.0&15 \\
                            \midrule
\multirow{3}{*}{Fusion}    
&\underline{Broad Sum}&67.7&39.9&68.0&29.7&14 \\
&Concat&66.8&38.3&66.7&27.2&14 \\
&W/o Prior&50.9&37.6&29.2&27.8&14\\
\midrule
\multirow{4}{*}{Single task}        
&\underline{Unification}&67.7&39.9&68.0&29.7&14 \\
&SOT only&67.5&-&-&-&14 \\
&MOT only&-&39.6&-&-&14 \\
&VOS only&-&-&68.4&-&14 \\
&MOTS only&-&-&-&28.1&14 \\
\midrule
\multirow{2}{*}{Speed}
&\underline{Ours}&67.7&39.9&68.0&29.7&14\\
&Ours-RT&67.1&37.5&66.8&26.2&23\\
\midrule
\end{tabular}
}}
\vspace{-4mm}
\label{table:Ablations}
\end{table}

\vspace{-3mm}
\section{Conclusions}

We propose Unicorn, a unified approach to address four tracking tasks using a single model with the same model parameters. For the first time, it achieves the unification of network architecture and learning paradigm for object tracking. Extensive experiments demonstrate that Unicorn performs on-par or better than task-specific counterparts on 8 challenging benchmarks. We hope that Unicorn can serve as a solid step towards the general vision model.

\noindent{\textbf{Acknowledgement. }
We would like to thank the reviewers for their insightful comments. Huchuan Lu and Dong Wang are supported in part by the National Natural Science Foundation of China under Grant nos. 62022021, 61806037, 61725202, U1903215 and 61829102, and in part by the Science and Technology Innovation Foundation of Dalian under Grant no.2020JJ26GX036 and Dalian Innovation leader’s support Plan under Grant no. 2018RD07. Ping Luo is supported by the General Research Fund of HK No.27208720, No.17212120, and No.17200622.}
%
%
\bibliographystyle{splncs04}
\bibliography{egbib}

\clearpage
\appendix

\section{Unified Head Architecture}
\begin{figure}[h]
  \begin{center}
\includegraphics[width=0.9\linewidth]{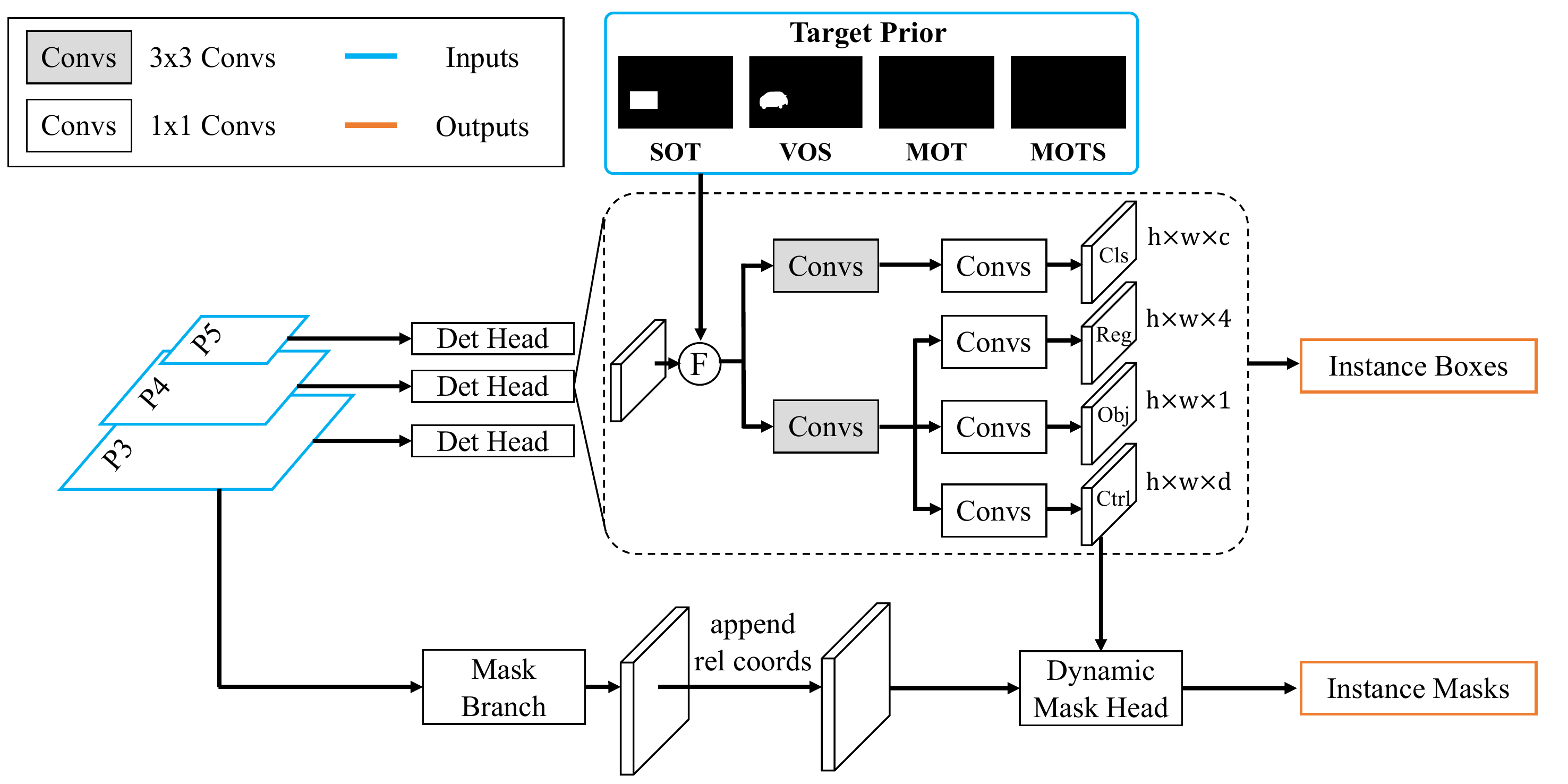}
  \end{center}
  \vspace{-5mm}
  \caption{Unified head architecture of the Unicorn.} 
  \label{fig-head}
\vspace{-5mm}
\end{figure}

The detailed head architecture is shown in Fig.~\ref{fig-head}. The unified head takes the original FPN feature $\mathbf{F}\in\mathbb{R}^{h\times{w}\times{c}}$ and the target prior $\mathbf{P}\in\mathbb{R}^{h\times{w}\times{1}}$ as the inputs. The two inputs are first fused by broadcast sum, getting the fused feature $\mathbf{F}^{'}\in\mathbb{R}^{h\times{w}\times{c}}$. Then the fused feature is passed to the detection head~\cite{YOLOX} and the instance segmentation head~\cite{CondInst}, predicting final boxes or masks. The head network is fully-convolutional, without any RoI operation such as RoI Align.

\section{Training Details}
Since accurate mask annotations are quite expensive while bounding boxes are relatively cheap, available training data of SOT\&MOT is usually dozens of times that of VOS\&MOTS. Directly mixing training data from four tasks will lead to a serious data-imbalance problem. To alleviate this problem, we divide the whole training process into two stages. Specifically, in the first stage, we randomly sampled training data from SOT datasets (COCO~\cite{COCO}, LaSOT~\cite{LaSOT}, GOT-10K~\cite{GOT10K} and TrackingNet~\cite{trackingnet}) and MOT datasets (For evaluating on MOT Challenge, we use Crowdhuman~\cite{crowdhuman}, ETHZ~\cite{ETHZ}, CityPerson~\cite{citypersons}, MOT17~\cite{MOT17}. For evaluating on BDD100K, we use the training set of BDD100K~\cite{BDD100K}) with a 1:1 sampling ratio to train the whole network without the mask head. In this stage, the network is optimized with the sum of the correspondence loss and the detection loss. $\mathbf{L}_{\mathrm{stage1}}=\mathbf{L}_{\mathrm{corr}}+\mathbf{L}_{\mathrm{det}}$. More details about $\mathbf{L}_{\mathrm{det}}$ can be found in the YOLOX paper.
Then in the second stage, to prevent the model from overfitting on the VOS\&MOTS and negatively impacting the performance of SOT\&MOT, we only train the mask head with the data from VOS (COCO~\cite{COCO}, DAVIS~\cite{DAVIS17}, Youtube-VOS~\cite{YoutubeVOS}) and MOTS (MOTS~\cite{MOTS}, BDD100K~\cite{BDD100K}), leaving other parameters fixed. $\mathbf{L}_{\mathrm{stage2}}=\mathbf{L}_{\mathrm{mask}}$
\vspace{-5mm}

\end{document}